\documentclass[runningheads]{llncs}
\usepackage[T1]{fontenc}
\usepackage{graphicx}
\usepackage{float}
\usepackage{wrapfig}
\usepackage{hyperref}
\usepackage{amsmath,amsfonts,graphicx}
\usepackage{subcaption}
\usepackage{tikz}
\usepackage{multirow}
\usetikzlibrary {shapes.geometric}
\usetikzlibrary {positioning}

\begin{document}

\title{Monitoring Time Series With Missing Values:\\a Deep Probabilistic Approach}

\author{Oshri Barazani\inst{1} \and David Tolpin\inst{2}}
\institute{PUB+ \and Ben-Gurion University of the Negev
\\ \email{\{oshribr,david.tolpin\}@gmail.com}}

\maketitle

\begin{abstract}
Systems are commonly monitored for health and security through
collection and streaming of multivariate time series. Advances
in time series forecasting due to adoption of multilayer
recurrent neural network architectures make it possible to
forecast in high-dimensional time series, and identify and
classify novelties early, based on subtle changes in the trends.
However, mainstream approaches to multi-variate time series
predictions do not handle well cases when the ongoing forecasts
must include uncertainty, nor they are robust to missing data.
We introduce a new architecture for time series monitoring based
on combination of state-of-the-art methods of forecasting in
high-dimensional time series with full probabilistic handling of
uncertainty. We demonstrate advantage of the architecture for
time series forecasting and novelty detection, in particular
with partially missing data, and empirically evaluate and
compare the architecture to state-of-the-art approaches on a
real-world data set.
\end{abstract}

\section{Introduction}

Modern information systems and operation environments are
commonly monitored through collection and streaming of
multivariate time series. The monitoring tasks comprise both
forecasting, for planning of resource allocation and decision
making, and novelty detection and characterization, for ensuring
faultless functioning and early mitigation of failures and
threats.  Advances in time series forecasting
due to adoption of multilayer recurrent neural network
architectures made it possible to forecast in high-dimensional
time series, and identify and classify novelties (anomalies)
early, based on subtle changes in the trends.  However,
mainstream approaches to multi-variate time series modelling
do not handle well cases when uncertainty is involved, either in
the input, when some of the observations are missing, or in the
output when the distribution of future observations, rather than
their point values, is predicted. For forecast uncertainty
modelling, stochastic latent variable
variants of high-dimensional time series models where
introduced, but so far have had to rely on sampling to account
for uncertainty, limiting the performance of data handling.
Imputation schemes were proposed for dealing with 
missing data, however, they do not generally give a satisfactory
solution in presence of transient unavailability of some of
the data sources (e.g. when a sensor stops working, or a
transport channel malfunctions), which is  a common case with
monitoring of complex systems.  

A systematic and theoretically founded approach to handling
both input and output uncertainty would thus constitute a
significant and welcome contribution to the theory and practice
of monitoring of multivariate time series. It would also be
highly desirable for such approach to facilitate efficient
offline (learning) and online (inference) computations. In this
ongoing research, we propose a deep learning architecture which
uses a simple but powerful extension of traditional recurrent
neural network (RNN) architecture which allows both 
\begin{itemize} 
\item to handle missing inputs
in some or all of the components in a multivariate time series,
\item and to accomplish multi-step probabilistic forecasting
\end{itemize}
in high-dimensional time series, paving a path to better
decision making and finer and more robust anomaly detection and
characterization.  We evaluate the architecture on a real-world
data set of multivariate time series collected from a 
network for cloud computing, and empirically demonstrate
advantage of the proposed architecture over commonly used
approaches.

\section{Problem: Multivariate Time Series Forecasting}

The core problem we address is forecasting in a multivariate
time series. Formally, a \textit{time series} is a matrix $X$
of shape $T\times N$, where $T$ is the number of time steps and
$N$ is the number of dimensions. The time steps are assumed to
be equispaced.  A $k$-step probabilistic \textit{forecast}
$\mathcal{F}_{tk}$ at time $t$ is the belief distribution of
time series $X_{t+1:t+k}$ for time steps $t+1 ... t+k$ given the
observed time series $X_{1:t}$ for time steps $1...t$.  

The forecasting is accomplished by applying model $\mathcal{M}_\theta$
parameterized by parameters $\theta$ to the observed time
series:
\begin{equation}
\mathcal{F}_{tk} = \mathcal{M}_\theta(X_{1:t})
\end{equation}
The machine learning task is to devise $\theta^*$
that gives the best forecast, in terms of a certain loss
function. A natural loss in the probabilistic setting is the
average negative log likelihood of $\theta$ given a training
data set $\mathcal{X}$ of multiple time series:
\begin{equation}
	\theta^* = \arg\min_\theta \mathbb{E}_{X \in \mathcal{X},t
	\in 1 ... T-k}\left[ -\log \Pr(X_{t+1:t+k}|M_\theta(X_{1:t}))\right]
	\label{eqn:theta-star}
\end{equation}
When the model is differentiable by $\theta$, the task is
usually accomplished by performing a stochastic gradient loss
minimization.

In the basic case, $X$ is real-valued, $X \in \mathbb{R}^{T
\times N}$. Here, we are interested in an extension of the basic
case, in which some of the elements can be missing from $X$, that is
$X \in (\mathbb{R} \cup \bot)^{T \times N}$.

\section{Architecture: Recurrent Neural Network with Uncertainty Propagation}

We introduce here a recurrent neural network architecture which
facilitates uncertainty propagation. The architecture is capable
both of handling missing values and of multi-step forecasting.
We begin with description of conventional forecasting with RNNs.
Then, we describe our proposed architecture as an extension to
the conventional model.

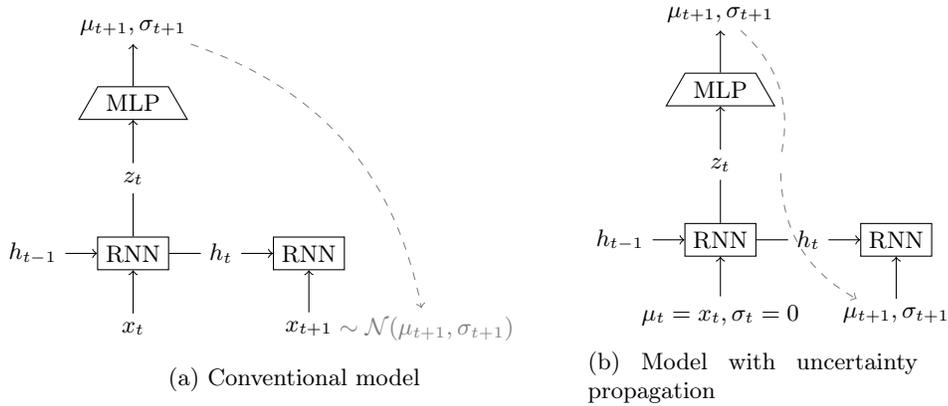
\begin{figure}
\begin{subfigure}{.64\textwidth}
\begin{tikzpicture}
\node [rectangle, draw] (rnn) {RNN};

\node (z_t) [above of=rnn] {$z_t$};
\node [trapezium, draw] (mlp) [above of=z_t] {MLP};
\draw [-] (rnn) to (z_t);
\draw [->] (z_t) to (mlp);

%output
\node (musigma) [above of=mlp] {$\mu_{t+1}, \sigma_{t+1}$};
\draw [->] (mlp) to (musigma);

%input
\node (x_t) [below of=rnn] {$x_t$};
\draw [->] (x_t) to (rnn);

\node (h_tm1) [left=12pt of rnn] {$h_{t-1}$};
\draw [->] (h_tm1) to (rnn);

\node (h_t) [right=12pt of rnn] {$h_{t}$};
\draw [-] (rnn) to (h_t);

\node (rnnnext) [rectangle, draw, right=12pt of h_t] {RNN};
\draw [->] (h_t) to (rnnnext);

\node (xnext) [below of=rnnnext] {$x_{t+1}$};
\draw [->] (xnext) to (rnnnext);

%sampling
\node (N) [gray,right=-4pt of xnext] {$\sim \mathcal{N}(\mu_{t+1}, \sigma_{t+1})$};
\draw [->, dashed,gray] (musigma) to [bend left] (N);

\end{tikzpicture}
\caption{Conventional model}
\label{fig:dt-conventional-model}
\end{subfigure}%
\begin{subfigure}{.36\textwidth}
\begin{tikzpicture}
\node [rectangle, draw] (rnn) {RNN};

\node (z_t) [above of=rnn] {$z_t$};
\node [trapezium, draw] (mlp) [above of=z_t] {MLP};
\draw [-] (rnn) to (z_t);
\draw [->] (z_t) to (mlp);

%output
\node (musigma) [above of=mlp] {$\mu_{t+1}, \sigma_{t+1}$};
\draw [->] (mlp) to (musigma);

%input
\node (x_t) [below of=rnn] {$\mu_t=x_t,\sigma_t=0$};
\draw [->] (x_t) to (rnn);

\node (h_tm1) [left=12pt of rnn] {$h_{t-1}$};
\draw [->] (h_tm1) to (rnn);

\node (h_t) [right=12pt of rnn] {$h_{t}$};
\draw [-] (rnn) to (h_t);

\node (rnnnext) [rectangle, draw, right=12pt of h_t] {RNN};
\draw [->] (h_t) to (rnnnext);

\node (xnext) [below of=rnnnext] {$\mu_{t+1}, \sigma_{t+1}$};
\draw [->] (xnext) to (rnnnext);

\node (F) [right=12pt of z_t] {};
\draw [-,dashed,gray] (musigma) to [bend left] (F);
\draw [->,dashed,gray] (F) to [bend right] (xnext);

\end{tikzpicture}
\caption{Model with uncertainty propagation}
\label{fig:dt-uncertainty-propagation}
\end{subfigure}
\caption{Time series models}
\label{fig:dt-time-series-models}
\end{figure}

\subsection{Conventional Forecasting}

A popular realization of the forecasting model
$\mathcal{M}_\theta$ is a recurrent neural network (RNN), with
$\theta$ corresponding to the network parameters.  There is a
range of neural recurrent models of varying complexity to deal
with time series forecasting. Most models include a recurrent
unit which threads the state through the time steps, accepts
data as inputs and produces next step predictions as outputs.
The simplest model is an RNN with a fully-connected readout
layer to produce forecasts
(Figure~\ref{fig:dt-conventional-model}).  RNN can be based on
LSTM~\cite{HS97}, GRU~\cite{CMB+14}, or another architectural variant, and
is often multi-layer.  Architectures  may also include
intermediate modules, and sampling-based variational
layers~\cite{CKD+15,YB21}. The overall architecture stays almost
the same, with more connections, intermediate modules and
sampling-based variational layers.

\paragraph{Input and Output} This architecture normally accepts
observation vectors and outputs vectors of distribution
parameters for the belief distribution of the observations at
the next time step. In the simplest case, the network produces a 
single output for each input, that is the dimensions of the
input and the output vector coincide. This corresponds to the
assumption of homoskedasticity of epistemic noise, and either
the mean squared error (corresponding to the Gaussian error
distribution) or the mean absolute error (corresponding to the
Laplace error distribution) is minimized. 

More generally though, the epistemic noise is better modelled
heteroskedastically, and a two-parameter loss distribution,
with the location and the scale as the parameters.
In the case of the frequently used normal (Gaussian)
distribution, the output vector consists of means $\mu$
(location) and standard deviations $\sigma$ (scale) of all
dimensions and is twice as wide as the input.

\paragraph{Training} The model is trained to maximize
probability of prediction. In the most basic case, called
\textit{out-of-sample one-step} forecasting, a single step
is predicted for each time step in the series. In an $n$-step
time series, steps $1 ... n-1$ are used as the input, and steps
$1 ... n$ as the ground truth. Following~\eqref{eqn:theta-star}, the
network is trained to minimize negative log probability of true
observations given the predicted belief distributions. More
generally, a model can also be trained to predict more than a
single step at once into the future, however this is rarely used in
practice because the necessary size of the training data set
grows exponentially with the prediction depth. Instead, future
predictions are produced recurrently during forecasting.

\paragraph{Forecasting} Forecasting is accomplished by passing
past observations through the model to obtain forecasts for the
future time steps. In the \textit{out-of-sample one-step} mode,
a single step into the future is forecast. If a longer forecast
is required, the current forecast is entered as the input at the
next time step, time after time, up to the required length.
Either the location (the point forecast) or a random sample from
the belief distribution is used as the future input. Using
random samples also allows to assess uncertainty multiple steps
into the future: one can repeatedly sample from the belief
distribution at each future step, and feed the sample as the
input to the following step. Then, based on produced samples at
future steps, one can estimate uncertainty intervals. Such
Monte-Carlo handling of uncertainty is quite expensive
computationally though, because the standard deviation of
prediction error decreases as slowly as $\sqrt{N}$
with the number of samples $N$, on one hand, and uncertainty
may, in general, grow exponentially with prediction depth, on
the other hand.

\paragraph{Novelty detection} Forecasts produced by the model
can be used for a number of purposes, including decision making
and, in particular, novelty (anomaly) detection. There are two
related but different phenomena indicating a novelty
in time series behavior:
\begin{enumerate}
\item Predicted volatility of the time series is high, that is, future
observations can only be forecast uncertainly (with high variance). 
\item Probability of actual observations, when observed, given a prediction
from a past state, is low.
\end{enumerate}
Either phenomenon, or both of them, can be used to alert about
novelties in the time series. In recurrent neural network
architectures, the hidden state ($h_t$ in
Figure~\ref{fig:dt-time-series-models}) can be used to identify
and classify anomalies.

\subsection{Forecasting with Uncertainty Propagation}

The basic scheme outlined above poses difficulty in
applications with high-dimensional time series and partially
missing observations. Sampling based uncertainty assertion
impacts performance, and missing observations are often imputed
heuristically~\cite{LKW16,SYG+19}. An architecture which
incorporates confidence about data and in which observed and
predicted data are interchangeable is  highly desirable. For
example, if out of 5 components 3 were measured and 2 predicted
from an earlier step we want to input all of them into the next
time step for further forecasting. In addition, the model
architecture should be capable of robust uncertainty prediction
and benefit from training with multiple steps of out-of-sample
data.

Our proposed architecture is based on the observation that if
(at least) the location and the scale are used  to represent
forecasts, an observation (that is, certain knowledge at a given
step) can also be expressed using two parameters, by setting the
location to the observation, and the scale to 0. For the normal
distribution $\mathcal{N}(\mu, \sigma)$, the location and scale
parameterization is straightforward, corresponding to $\mu$ and
$\sigma$, however other belief distributions can be parameterized
by location and scale as easily, e.g. the log-normal,
Gamma, or Laplace distribution. For conciseness, we will confine
further discussion to the case of independent normal belief
distributions for each component; however, other distribution
shapes can also be used. Based on this observation,
\textbf{we propose the following
extension} to the conventional RNN-based forecasting model
(Figure~\ref{fig:dt-uncertainty-propagation}):

\begin{enumerate}
\item \textit{The input, as well as the output, is a vector of
distribution parameters}. For the independent normal
distributions, the distribution parameter vector consists of the
means followed by the standard deviations. If the data has 5
components, the input will be 10-dimensional. For observed data
--- measurements present at the current time step --- the
standard deviation is zero. For missing data the input is the
mean and the standard deviation as predicted from the preceding
time steps.

\item  Training can, in principle, be accomplished on data with
missing values, but training on data with missing values
incurs performance drawbacks and should be avoided. First,
handling missing values and replacing them with early
predictions introduces contingency in the forward run of the RNN
and slows down significantly the execution during training.
Second, missing values should, in general, themselves be viewed
as anomalies. One must be able to handle them during inference, 
but should not rely on their presence in the training data.

Therefore, we devise a scheme for \textit{training our model on data
that does not contain missing values}. Even in applications where
missing values are common in inference, training data without
missing values is usually readily available. However,
since we introduce confidence into the input, we cannot
train the network myopically, in out-of-sample one-step manner --- 
the standard deviations in the input data will always be zero,
and the network will never learn how to use them. To overcome
this, we train on multiple predicted steps. We feed each
prediction, without sampling, as input to the next step and
compute the loss as negative log probability of this number of
future points versus our prediction.
\end{enumerate}

To illustrate, given the data set of 5 dimensions, the input has 10
dimensions. If we train with 3 time steps lookahead, the ground
truth will be a matrix of size $3\times 5$. The prediction against
which the likelihood of this ground truth is computed will be
a matrix of size $3 \times 10$. Intuitively, we would expect the
predicted standard deviation to increase along the time axis
for each component.

The ability of probabilistic forecasting with uncertainty, in
the form of multivariate normal distributions, far into the
future, opens opportunity for application to more robust novelty
detection approaches. Instead of detecting novelty based on log
probability of observations given predictions from the
past~\cite{CBK09}, which is prone to false positives due to
observation noise, novelties can be detected and analysed by
comparing predictions of the same time point from different
points in the past.  In this case,
KL-divergence between predictions provides a theoretically sound
and robust mechanism for detection of anomalies, and is in
particular relevant for monitoring of large operation
environments with high dimensionality of time series and
occasional missing values and heteroskedastic noise~\cite{ASH08,T19}.

\section{Case Study: Monitoring a Computer Cloud}

We evaluate the proposed architecture on a data set of
monitoring a cluster of 100 computing nodes in the cloud. For
each node, the incoming and the outgoing network traffic (in
bytes) and the CPU usage (relative) are logged with 1 minute
resolution. 240 hours were logged, resulting in 12000 120-minute
3-dimensional samples.  We split the dataset into the training,
validation, and test as 80\%, 10\%, and 10\% correspondingly.
Since the original data set does not have many missing data
points, we emulated data sets with missing data by randomly
removing 5\%, 10\%, 20\%, and 50\% of the data.

We used a 3-layer GRU-based recurrent neural network with hidden
size 64 and 20\% dropout between layers. We trained the network
with lookahead depths (number of steps to forecast in the
future) 2, 4, 8, and 16 using the Adam optimizer with learning
rate 0.001, training for 20 epochs (sufficient for convergence).
We performed the training on a cloud computing node with 1
NVIDIA T4 GPU, 4 Intel Xeon Platinum CPUs, and 64 Gb memory. The
training of a single model took 20 minutes.

\begin{table}
\caption{Uncertainty propagation vs. `replace by the mean`.}
\label{tab:replace-by-mean}
	\setlength\tabcolsep{18pt}
	\centering
	\begin{tabular}{r|c c c c}
	 \textbf{missing} & \textbf{2} & \textbf{4} & \textbf{8} & \textbf{16} \\ \hline
		  \textbf{5\%} & 0.001 & 0.001 &  0.06 & 0.10  \\
		 \textbf{10\%} & 0.001 & 0.002 &  0.08 & 0.11  \\
		 \textbf{20\%} & 0.003 & 0.003 &  0.11 & 0.13  \\
		 \textbf{50\%} & 0.004 & 0.006 &  0.12 & 0.16  
	\end{tabular}
\end{table}
\begin{table}
\caption{Uncertainty propagation vs. `replace by a random sample`.}
\vspace{1em}
\label{tab:replace-by-sample}
	\setlength\tabcolsep{18pt}
	\centering
	\begin{tabular}{r|c c c c}
	 \textbf{missing} & \textbf{2} & \textbf{4} & \textbf{8} & \textbf{16} \\ \hline
		  \textbf{5\%} & 0.04 & 0.04 & 0.13 & 0.14 \\
		 \textbf{10\%} & 0.06 & 0.06 & 0.16 & 0.24 \\
		 \textbf{20\%} & 0.11 & 0.12 & 0.20 & 0.27 \\
		 \textbf{50\%} & 0.18 & 0.19 & 0.28 & 0.30   
	\end{tabular}
\end{table}
We compared our approach with conventional imputation methods
`replace by the mean' and `replace by a random sample'.
In the `replace by the mean' method, a missing value is replaced 
by the mean of the forecast. In the `replace by a random
sample' method, a missing value is replaced by a random sample
drawn from the forecast. As a performance metrics, we
used per-point negative log-likelihood loss on the test set. 
Tables~\ref{tab:replace-by-mean} and~\ref{tab:replace-by-sample}
show the difference in loss between uncertainty propagation and
`replace by the mean' and `replace by a random sample',
correspondingly. The greater is the number, the worse is the
forecasting by each of the methods compared to uncertainty
propagation.
One can see that in all cases uncertainty propagation provides
better forecasts than either of the conventional methods. 

\begin{figure}
\centering
\includegraphics[width=0.95\linewidth]{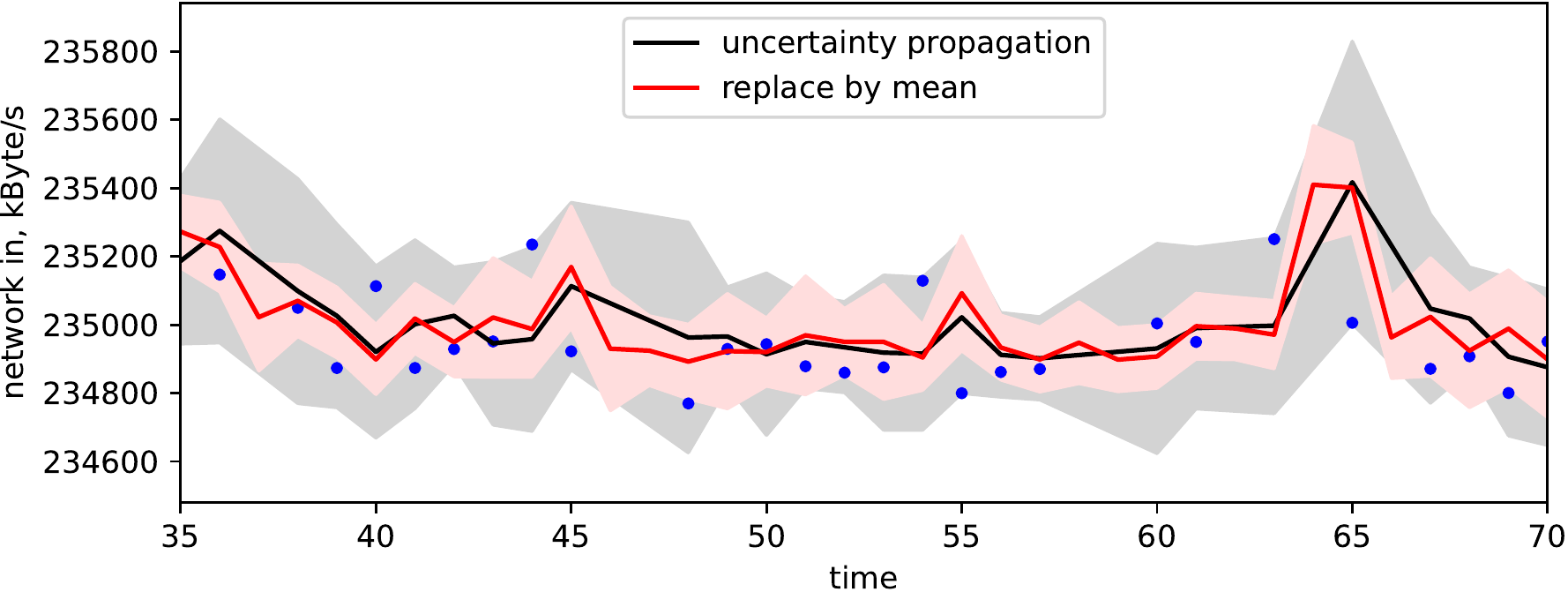}
\caption{Uncertainty propagation vs `replace by the mean'. 95\% confidence intervals are shaded.}
\label{fig:dist-vs-mean}
\end{figure}
As an illustration of the advantage of uncertainty propagation,
consider Figure~\ref{fig:dist-vs-mean}, which shows forecasts
using uncertainty propagation and `replace by the mean' in presence
of missing values. Forecasts through uncertainty propagation
result in adequate confidence intervals. However, when missing
values are replaced by the mean of the belief distribution,
further forecasts are overconfident and too many observations
fall outside of 95\% confidence intervals.

The code and data for the case studies are available at
\url{https://bitbucket.org/dtolpin/dbts-studies/}.

\section{Related Work} 

There appear to be two interconnected areas related to this
research. One area is uncertainty representation and propagation
in recurrent neural models. The other area is handling of
missing values in time series, again in the context of
recurrent neural models in particular. 

The importance of uncertainty quantification in deep learning is
well understood~\cite{MFS+21}.  Recurrent neural networks can
express forecast uncertainty through predicting distribution
parameters, such as the mean and the standard deviation, instead
of point values~\cite{HS97}. When expressing uncertainty by closed-form
distributions is insufficient, stochastic latent
variables are introduced into RNNs~\cite{YB21,CKD+15,FSK+16}. 
Uncertainty representation in RNNs is related to uncertainty
propagation and multi-step forecasting. For multi-step
forecasting, uncertainty must be propagated multiple steps into
the future. Uncertainty propagation is usually achieved through
random sampling during training or inference~\cite{LYY+19,AV20,YB21}. 
Our approach differs in that conventional RNN architectures are
leveraged to represent uncertainty in both the input and
the output, and that uncertainty propagation is accomplished
deterministically, without resorting to random sampling, which
facilitates efficient training and inference.

Handling of missing values in time series has inspired research
for decades due to the fact that many otherwise efficient and
robust algorithms, in particular those based on recurrent neural
architectures, require that all values in the time series
are present and lie within a valid range~\cite{WSY+21}. A
widespread approach is to \textit{impute} the data,
that is, to replace missing values with values inferred from
other values in the same time series or in other time series
in the data set~\cite{KC18,SYG+19}. Alternatively, a
missing value is treated as an observation itself, often by
introducing an auxiliary indicator variable~\cite{LKW16,BDS21}.
In our work, we take a third approach --- a missing value,
either due to an absent observation or in the course of
multi-step forecasting, is replaced by a parametrically
specified belief distribution of the value based on the past
observations. 

\section{Discussion and Future Research}

We presented a deep probabilistic architecture for uncertainty
propagation in multivariate time series. This architecture
organically handles two important problems in deep time series
modelling: missing data and multi-step forecasting. Empirical
evaluation demonstrated that our approach outperforms
conventional baselines in terms of forecasting accuracy, while
still being easy to implement. Since, unlike some other
approaches to uncertainty propagation, our architecture
avoids sampling, uncertainty can be propagated efficiently and
represented in closed parametric form, rather than approximated
by samples and posterior intervals.

We confined most of the discussion to the normal uncertainty shape.
Other distributions can be used instead of the normal
distributions where appropriate, provided their parameterization
allows to express a certain observation as well as an uncertain
belief. Analysis of distributions for representing uncertainty
and their feasible parameterization is a subject of ongoing
research. Another research direction worth exploring is
extension of the presented architecture to bidirectional
recurrent neural networks~\cite{BRH15}. Bidirectional RNNs allow
to account for both past and future observations where
appropriate, but apparently make uncertainty propagation
more complicated. Still, preliminary results suggest that
uncertainty in bidirectional RNNs can be handled in a similar
manner, further facilitating efficient probabilistic uncertainty
propagation in a broader class of deep learning models for time
series.

\section*{Acknowledgements}

We thank PUB+ for providing computational facilities for
conducting the empirical evaluation. David Tolpin is partially
supported by Israel-U.S. Industrial Research and Development
Foundation's \textit{Cybersecurity technology for critical power
infrastructure AI-based centralized defense and edge resilience}
project.

\nocite{BDS21}
\nocite{CPK+18}
\nocite{SYG+19}
\nocite{KC18}
\nocite{LKW16}
\nocite{BRH15}
\nocite{AV20}
\nocite{FSK+16}
\nocite{MFS+21}
\nocite{FSP+16}
\nocite{SZN+19}
\nocite{LYY+19}
\nocite{CYP+21}
\nocite{YB21}
\nocite{CKD+15}
\nocite{HS97}
\nocite{CMB+14}
\nocite{WSY+21}

\bibliographystyle{splncs04}
\bibliography{refs.bib}

\end{document}